\documentclass[10pt,twocolumn,letterpaper]{article}

\usepackage{iccv}
\usepackage{times}
\usepackage{epsfig}
\usepackage{graphicx}
\usepackage{amsmath}
\usepackage{amssymb}

\usepackage[breaklinks=true,bookmarks=false]{hyperref}

\usepackage{microtype}
\usepackage{subfigure}
\usepackage{booktabs} 

\usepackage{algorithm}
\usepackage{algorithmic}

\usepackage{multirow}
\newcommand{\R}{\mathbb{R}}

\DeclareMathOperator*{\mha}{MHA}
\DeclareMathOperator*{\AP}{AP}
\DeclareMathOperator*{\ffn}{FFN}
\DeclareMathOperator*{\LN}{LN}
\DeclareMathOperator*{\KL}{KL}

\iccvfinalcopy 

\def\iccvPaperID{1234} 

\ificcvfinal\fi

\begin{document}

\title{KS-DETR: Knowledge Sharing in Attention Learning for Detection Transformer}

\author{Kaikai Zhao\\
Toyota Technological Institute, Japan\\
{\tt\small zhaokaikai@toyota-ti.ac.jp}
\and
Norimichi Ukita\\
Toyota Technological Institute, Japan\\
{\tt\small ukita@toyota-ti.ac.jp}
}

\maketitle
\ificcvfinal\thispagestyle{empty}\fi

\begin{abstract}  
Scaled dot-product attention applies a softmax function on the scaled dot-product of queries and keys to calculate weights and then multiplies the weights and values.
In this work, we study how to improve the learning of scaled dot-product attention to improve the accuracy of DETR. Our method is based on the following observations: using ground truth foreground-background mask (GT Fg-Bg Mask) as additional cues in the weights/values learning enables learning much better weights/values; with better weights/values, better values/weights can be learned. We propose a triple-attention module in which the first attention is a plain scaled dot-product attention, the second/third attention generates high-quality weights/values (with the assistance of GT Fg-Bg Mask) and shares the values/weights with the first attention to improve the quality of values/weights. 
The second and third attentions are removed during inference. We call our method knowledge-sharing DETR (KS-DETR), which is an extension of knowledge distillation (KD) in the way that the improved weights and values of the teachers (the second and third attentions) are directly shared, instead of mimicked, by the student (the first attention) to enable more efficient knowledge transfer from the teachers to the student. Experiments on various DETR-like methods show consistent improvements over the baseline methods on the MS COCO benchmark. 
Code is available at \url{https://github.com/edocanonymous/KS-DETR}.

 
\end{abstract}

\section{Introduction}
\label{sec:intro}

\begin{figure}[t!]
\centering
\subfigure{\includegraphics[width=0.46\textwidth]{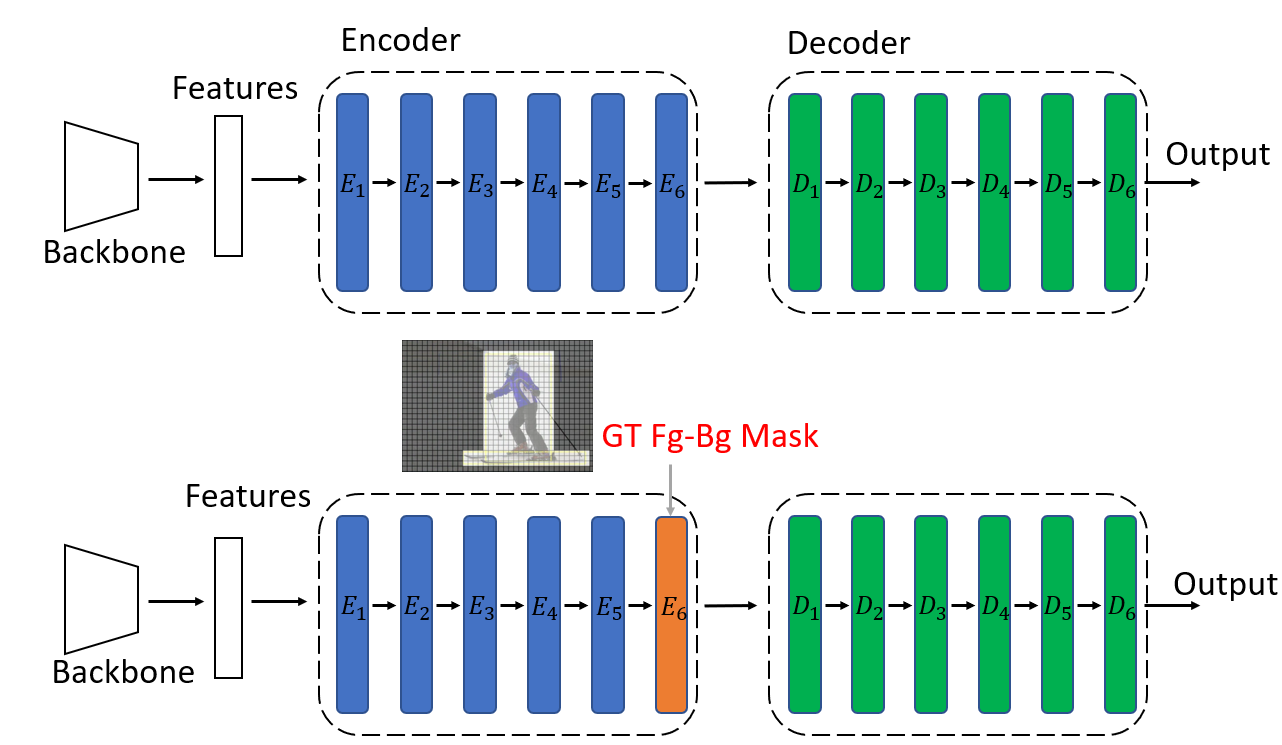}}
\caption{The DETR architecture (top) and our KS-DETR architecture (bottom). We replace the scaled dot-product attention of DETR with our triple-attention in the last encoder layer. Note that the position encoding in the encoder and decoder, and learnable object queries in the decoder are skipped in the architectures for clarity.}
\label{fig:KS-DETR-framework}
\end{figure}


 Detection transformer (DETR \cite{carion2020end}), built on a transformer encoder-decoder architecture, greatly simplified the object detection pipeline of traditional object detection methods. It views object detection as a set prediction problem by bipartite matching to enforce unique predictions and outputs a fixed number of object classes and locations given a fixed set of learnable object queries.

 Scaled dot-product attention, applied in the self-attention module in the encoder/decoder layers and cross-attention module in the decoder layers, is an essential component for DETR. Given input features $X$, it first conducts linear projections on $X$ to obtain queries $Q$, keys $K$ and values $V$. Then it applies a softmax function on the scaled dot-product of $Q$ and $K$ to calculate weights (or attention map $A$) and then multiplies $A$ and $V$ to obtain the output features $Y$. 
  
 DETR-like methods have improved the attention learning of DETR significantly by, for instance, using multi-scale features with deformable attention \cite{zhu2020deformable}, decoupling the attention to content attention and spatial attention \cite{meng2021conditional}, improving the design of the learnable object queries \cite{wang2022anchor,liu2022dab,li2022dn}. In this paper, we provide a new perspective on improving attention learning.  Our work is closely related to attention distillation.
 

 Attention distillation \cite{rubin2021attention,jiao2019tinybert,sun2019mobilebert,wang2020minilm}, as an application of knowledge distillation (KD \cite{bucilua2006model,hinton2015distilling}), has been used to improve attention learning to force a student model to mimic the attention maps of a teacher model.
 The attention distillation loss between the attention maps of the teacher $A^T$ and the student $A^S$ is typically defined by

 \begin{equation}\label{eq:attention-distill-loss} 
 L = \frac{1}{H} \sum_{h=1}^{H}\KL(A_{h}^{T} \| A_{h}^{S}),
\end{equation}

\noindent where $\KL$ represents the Kullback-Leibler divergence loss and $H$ is the number of attention maps.
 The motivation for conducting attention distillation is that if the student can learn better attention maps by mimicking the teacher's attention map, it is also likely to learn better output features $Y$.

\begin{figure*}[t!]
\centering
\subfigure{\includegraphics[width=0.90\textwidth]{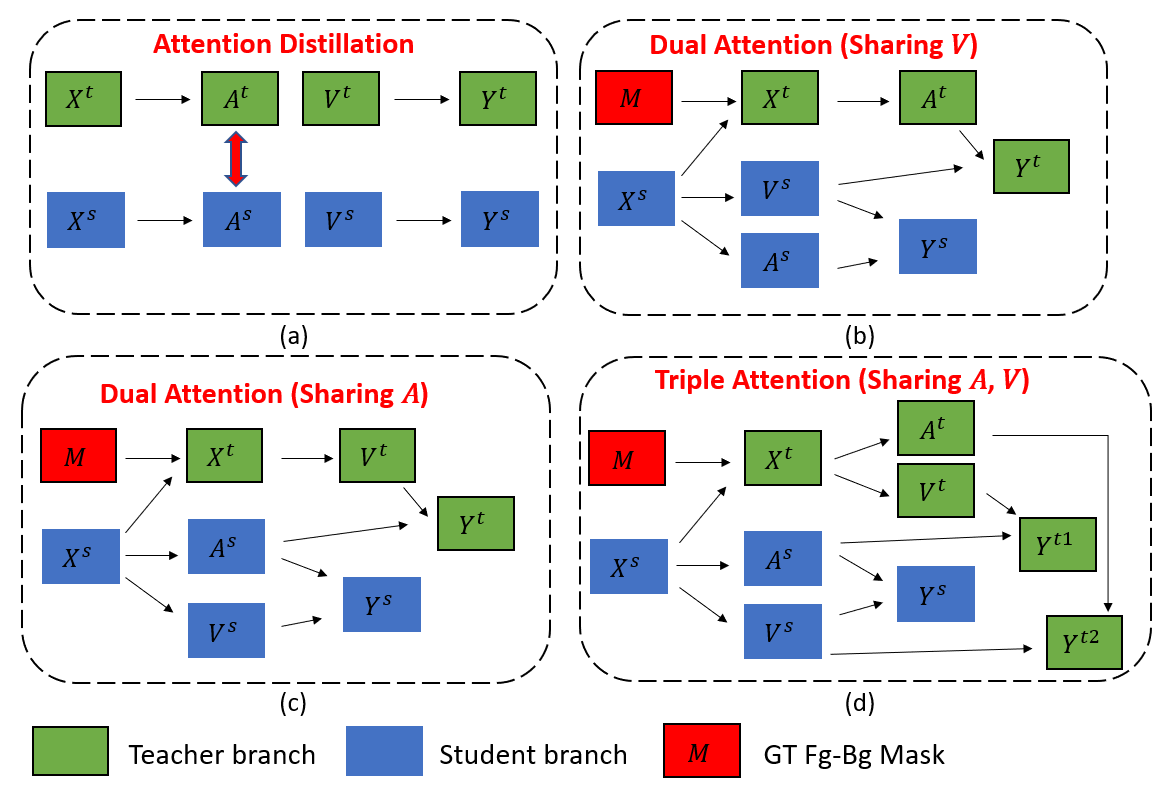}}
\caption{Difference between attention distillation and our knowledge (attention maps $A$ and values $V$) sharing framework with triple-attention. 
Dual-attentions in (b) and (c) are two variants of our triple-attention, obtained by removing a different teacher-attention module each time from our triple-attention.
Attention distillation requires training a large teacher model to provide the teacher attention map $A^t$, while our method learns $A^t$ inside the student model by using GT Fg-Bg Mask $M$ as additional cues. 
Here $X^{s}$ is the input feature of the scaled dot-product attention of the student attention. We first obtain the teacher feature $X^t$ by fusing $X^{s}$ with $M$ (details are given in Sec. \ref{sec:ks-detr}). Then we derive $A^t$ and teacher values $V^t$ from $X^t$. 
Note that $Y^{t1}$, $Y^{t2}$ in (d) are the outputs of the scaled dot-production of our two teacher attentions.
}
\label{fig:knowledge-sharing-framework}
\end{figure*}

 However, there are two issues for attention distillation. First, a large trained teacher model is needed to provide the teacher attention map, while training a large teacher model is time-consuming.  
Second, attention distillation ignores the gap in the representation ability between the teacher values and student values. The improvement in the output features can be difficult to achieve if the student has low-quality  values due to low model capability, even if the student successfully learns the essential part of the attention maps of the teacher. 

To address the first issue, we propose to use the ground truth foreground-background mask (GT Fg-Bg Mask) as additional cues to learn good teacher attention and values inside the student model. The GT Fg-Bg Mask is obtained by assigning 1 to pixels inside a ground truth bounding box and 0 otherwise. In object detection tasks, we need to decide the class label and exact boundaries for a predicted object. 
The GT Fg-Bg Mask will assist the attention learning for both the localization and classification tasks as it clearly identifies which pixel is foreground/background. 
With the GT Fg-Bg Mask, we can obtain the teacher attention by building separate branches which learn student attention and teacher attention separately, as shown in Fig. \ref{fig:knowledge-sharing-framework} b, c and d.

 For the second issue, we propose a knowledge-sharing strategy for more efficient knowledge transfer in attention learning. Our proposal is based on the following observations. The quality of $A$/$V$ improves as $V$/$A$ improves as one has to adapt the other toward learning good output features through backpropagation. Based on these observations, we design a triple-attention module in which the first attention is a plain scaled dot-product attention, the second attention generates high-quality $A$ (with the assistance of GT Fg-Bg Mask) and shares $V$ with the first attention to improve the quality of $V$, and the third attention generates high-quality $V$ and shares $A$ with the first attention to improve the quality of $A$. 

The differences between attention distillation and our knowledge sharing are shown in Fig. \ref{fig:knowledge-sharing-framework}.
The knowledge (attention maps and values) learned in the teacher-attention in our method is directly shared, instead of mimicked as in attention distillation, by the student. As a result, the gap in the representation ability of $V$ between the teacher and student is eliminated.

Our contributions are as follows:
\begin{itemize}
\item We show that high-quality attention maps and values can be learned with the assistance of the GT Fg-Bg Mask without training a large teacher model. 
\item We propose a triple-attention module to exploit the high-quality attention maps and values to improve the learning of scaled dot-product attention and obtained consistent improvements over various DETR-like methods.
\end{itemize}

\section{Related Work}
\label{sec:related-work}

 \subsection{DETR methods}

Followups of DETR \cite{carion2020end} have improved the attention learning of DETR significantly.
Deformable DETR \cite{zhu2020deformable} replaces the global dense attention with deformable one and limits the number of keys the query can attend to by sparsely sampling the key points.
SMCA \cite{gao2021fast} constraints the cross-attention to focus more on locations that are likely to contain objects by using a Gaussian-weighted spatial map for predicted object centers and scales.
DAB-DETR \cite{liu2022dab} uses box coordinates as queries in the decoder and restricts the regions of interest for cross-attention learning.  
Conditional DETR \cite{meng2021conditional} decouples the cross attention in the decoder to content attention and spatial attention. 
Our work is complementary to the above methods as our method adds two teacher attentions to assist the learning of normal attention.

\subsection{KD and attention distillation}

KD \cite{bucilua2006model,hinton2015distilling} was first introduced to compress the knowledge of a large teacher model to a small student model for classification tasks. The student is forced to not only predict the normal hard labels but also mimic the predicted category probability (soft labels) of the teacher, as the soft labels contain rich information that could not be encoded in the hard labels, e.g., the similarity of the output categories. Later works have extended KD to mimic, for instance, features and attention maps \cite{gou2021knowledge}. 

Early works of attention distillation for transformer architectures are from natural language processing. 
TinyBERT \cite{jiao2019tinybert} applies $\ell_{2}$ attention transfer loss to capture the attention knowledge from the teacher (BERT \cite{devlin2018bert}) to the student (TinyBERT).
MobileBERT \cite{sun2019mobilebert} minimizes the KL-divergence between the attention maps of the teacher and student.
MiniLM \cite{wang2020minilm} proposes to mimic not only the distribution of the attention maps (i.e., the scaled dot-product of queries-keys) but also the relation between values (i.e., the scaled dot-product of values-values). However, MiniLM requires the student and teacher to have the same number of attention heads for distribution mimicking. To solve this problem, MiniLMv2 \cite{wang2020minilmv2} proposes to completely replace the distribution mimicking with relation mimicking for the queries and keys. 
AttnDistill \cite{wang2022attention} applies attention distillation in a self-supervised vision transformer for classification tasks. It conducts interpolation and aggregation when the student and teacher have different numbers of attention heads and sizes of attention maps, respectively.
The most related work to us is \cite{rubin2021attention}. It conducts attention distillation for detection transformers. The attention maps of the last encoder layer from a large teacher network are distilled to the same location of a student network with fewer encoder and decoder layers. 
However, all the above methods require a large pretrained teacher model and ignore the representation gap of the values as we discussed in Sec. \ref{sec:intro}. 
We address these two issues in our method. Moreover, the attention maps of student and teacher in our method always have the same dimensions (cf. Sec. \ref{sec:ks-detr}) and thus our method is more general than attention distillation.

\subsection{Attention or feature learning with GT}

Ground truth (GT) is typically used as the target in supervision tasks to assist the training of a model by imposing a loss on the predictions of the model to be the same as the target.
In anchor-based two-stage detection methods such as Faster R-CNN \cite{ren2015faster}, one trick to improve the accuracy is to directly add GT boxes as high-quality proposals to increase the diversity of the input proposals in the second stage for better feature learning. The high-quality proposals are typically difficult to obtain if we only rely on the proposals made in the first stage.
DN-DETR \cite{li2022dn} and DINO-DETR \cite{zhang2022dino} adopted similar ideas to DETR, they add GT boxes (with random noise in object boundary and class label) as object queries to increase the learning of the decoder features. 
Our method differs from the above two methods in the way that we directly use GT boxes as additional features, instead of box priors, to improve attention learning and value learning.

\section{Our Method}
\label{sec:our-method}

We first review DETR \cite{carion2020end} and then introduce how we build our method upon DETR.

\subsection{DETR}

\begin{figure*}[t!]
\centering
\subfigure{\includegraphics[width=0.8\textwidth]{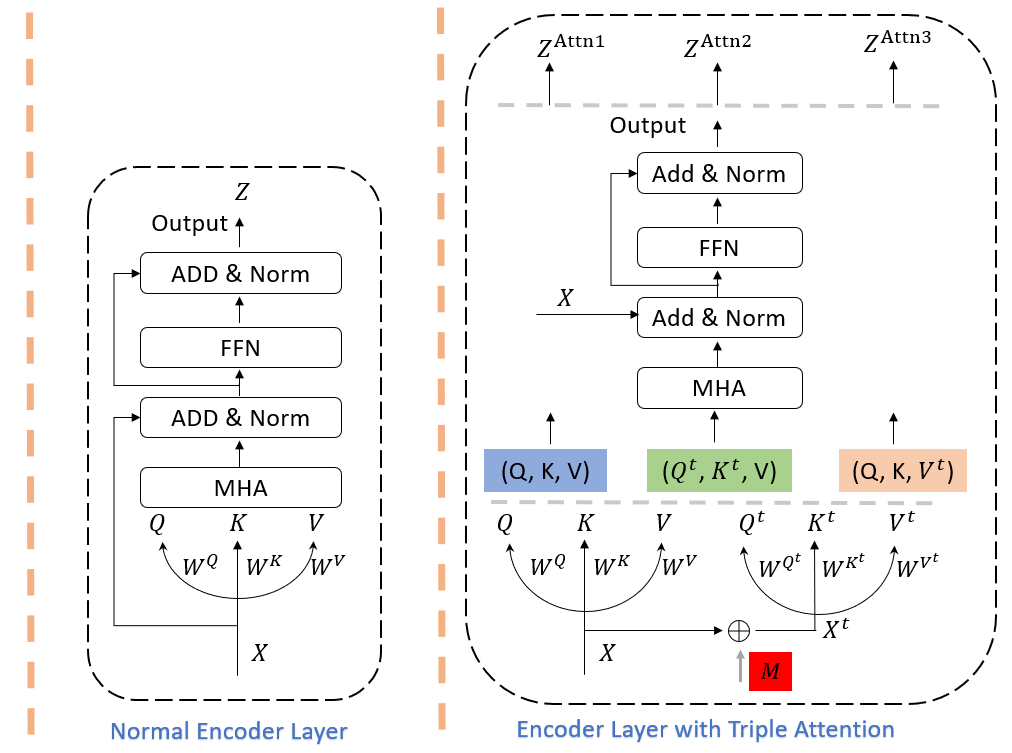}}
\caption{Details of the triple-attention used in our method (right) and the normal scaled dot-product attention (left). The $M$ in the red block represents the GT Fg-Bg Mask.}
\label{fig:triple-attention-module}
\end{figure*}


The DETR \cite{carion2020end} architecture consists of a (CNN) backbone, an encoder-decoder transformer, and a feed-forward network (FFN) as shown at the top of Fig. \ref{fig:KS-DETR-framework}.
The backbone generates features $f \in \R ^{H \times W \times C}$ from input image $I \in \R ^{H_0 \times W_0 \times 3}$. Before these features are fed to the encoder, they are first passed to $1 \times 1$ convolution to reduce the channel dimension from $C$ to $d$ and then flattened to tokens $X \in \R ^{HW \times d }$ along the spatial dimension. $X$ is further processed by a sequence of identical encoder layers  in the encoder to obtain encoder features $f^{e} \in \R ^{HW \times d}$ and a sequence of identical decoder layers in the decoder to obtain decoder features $f^{d} \in \R ^{N \times d}$ with respect to $N$ object queries. The object queries are learnable input embedding and $N$ is a hyperparameter of DETR. Finally, the FFN predicts the box coordinates and class label for each object query. 

Each encoder layer consists of a self-attention and position-wise FFN.
Without loss of generality, the forward propagation of the self-attention and FFN are given by 

\begin{equation}
\begin{aligned}
Z^{'} &= X + \LN(\mha(Q, K, V)) \\ 
 Z &= Z^{'} + \LN(\ffn(Z^{'})),
 \label{eq:encoder-forward}
\end{aligned}
\end{equation}

\noindent where $Q$, $K$ and $V$ are learned from $X$ and are the query, key and value, respectively, $\LN$ indicates layer normalization \cite{ba2016layer} and $\mha$ indicates multi-head attention \cite{vaswani2017attention}.
$\mha$ splits the input tokens $X$ into $h$ groups $X_1, ..., X_h$ (e.g., $h = 8$) along the channel dimension, conducts scaled dot-product attention on each group separately and then applies a linear projection to the outputs of $h$ heads to generate the final output. 
The linear projection is given by 

\begin{equation}\label{eq:out-proj} 
\mha(Q, K, V) = Concat(head_1, ..., head_h) W^{O},
\end{equation}

\noindent where $head_i = Attention(X_i W_i^{Q}, X_i W_i^{K}, X_i W_i^{V})$ represents the output of a single head and is estimated by

\begin{equation}\label{eq:attn} 
Attnention(Q, K, V) = softmax \left( \frac{Q K^{T}}{\sqrt{d_k}} \right) V.
\end{equation}

\noindent Here $softmax \left( \frac{Q K^{T}}{\sqrt{d_k}} \right)$ is often referred to as the attention map $A$.
As the transformer architecture is permutation-invariant, positional encoding ($PE$) is added to $X$ to obtain $Q$, $K$ and $V$ in each attention layer, i.e., 

\begin{equation}\label{eq:student-qk}
[Q; K; V] = [(X + PE) W^{Q}; (X + PE) W^{K}; X W^{V}]
\end{equation}

\noindent where $W^{Q} \in \R ^{d_{model} \times d_q}, W^{K} \in \R ^{d_{model} \times d_k}$, and $W^{V} \in \R ^{d_{model} \times d_v}$ are the parameters of the scaled dot-product attention, and $d_{model} = d / h$.


\subsection{KS-DETR}
\label{sec:ks-detr}



Our motivation is to improve the output features $Y$ of the scaled dot-product attention by improving both the attention maps $A$ and values $V$ by knowledge-sharing. Our hypothesis is that if $A$/$V$ is replaced with a high-quality one (i.e., $A^t$/$V^t$), then $V$/$A$ should also be improved as they have to adapt themselves to fit $A^t$/$V^t$ through backpropagation. As a result, the improved $V$/$A$ can be directly exploited by the normal scaled dot-product attention.



We verify the effectiveness of our knowledge-sharing idea in the encoder due to its simplicity. The encoder layer does not need to learn the encoder-decoder attention (or cross-attention) as in decoder layers.
Our method is designed to be general for DETR-like methods that use an encoder-decoder architecture. The framework of KS-DETR is shown at the bottom of Fig. \ref{fig:KS-DETR-framework}.
We replace the scaled dot-product attention in the last encoder layer with our triple-attention with the assistance GT Fg-Bg Mask.

\noindent\textbf{GT Fg-Bg Mask generation.} We first separate the foreground and background in the image space by
\begin{equation}\label{eq:dist-temporal}
\begin{matrix}
M_{I}(i,j) =  \begin{cases}
1 & \text{if } (i,j) \in \text{GT Boxes}  \\
0 & \text{otherwise}
\end{cases}
\end{matrix}
\end{equation}

\noindent where $i$ and $j$ represent the image coordinates on the horizontal and vertical directions, respectively, and GT Boxes represent the ground truth bounding boxes.
We then bilinearly interpolate $M_{I}$ to the size of the output feature maps of the backbone. Finally, we flatten the interpolated mask along the spatial dimension and obtain the binary mask $M \in \R ^{HW \times 1}$. 

\noindent\textbf{Triple-attention.} Our triple-attention module consists of a student attention and two teacher attentions.
It is derived from three groups of ($Q, K, V$) with repeated elements as shown in Fig. \ref{fig:triple-attention-module}. The three groups are: 1) ($Q, K, V$) for the first plain attention $Attn1$, 2) ($Q, K, V^{t}$) for the second attention $Attn2$ ($Q$ and $K$ are shared with $Attn1$), and 3) ($Q^{t}, K^{t}, V$) for the third attention $Attn3$ ($V$ is shared with $Attn1$). Here the superscript $t$ indicates \emph{teacher}, a term borrowed from knowledge distillation.
Note that there are other combinations of ($Q, K, V$), for instance, ($Q^{t}, K^{t}, V^{t}$), ($Q^{t}, K, V$) and ($Q, K^{t}, V$). However, ($Q^{t}, K^{t}, V^{t}$) cannot share $A$ or $V$ with $Attn1$, and ($Q^{t}, K^{t}, V$) for $Attn3$ is favored over ($Q^{t}, K, V$) and ($Q, K^{t}, V$) because both $Q^{t}$ and $K^{t}$ contribute to the learning of high-quality $A^t$.

As with the student model that is equal to the original DETR shown in Eq. \ref{eq:student-qk}, $Q^{t}, K^{t}, V^{t}$ are obtained from the linear projections of the teacher feature $X^{t}$ (explained below) by

\begin{equation}
\begin{aligned}
[Q^{t}; K^{t}; V^{t}] = [& (X^{t} + PE) W^{Q^{t}};  \\ 
& (X^{t} + PE) W^{K^{t}}; X^{t} W^{V^{t}}].
\label{eq:teacher-qk} 
\end{aligned}
\end{equation}

\noindent\textbf{Teacher feature generation.} The teacher feature $X^{t}$ is the output of the fusion of the input token $X$ and the GT Fg-Bg Mask $M$.
We could generate $X^{t}$ by simply concatenating $X$ and $M$ alone the channel dimension 

\begin{equation}\label{eq:teacher-x-original} 
X^{t} = concat(X, M), X \in \R ^{HW \times d}, M \in \R^{HW \times 1}.
\end{equation}

\noindent However, the concatenation will change the feature dimension and cause the resulting number of channels to be non-divisible when calculating multi-head attention. 


Here we propose to conduct sparse MLP (sMLP) that applies a positional-wise linear projection followed by a ReLU activation to only foreground tokens by 

\begin{equation}
\begin{aligned}
 X^{t}  &=  sMLP(X, M)   \\
 &=  X \odot (1 - M) + Relu(X W^{X}) \odot M, 
\label{eq:smlp} 
\end{aligned}
\end{equation}

\noindent where $W^{X} \in d \times d$ represents the parameters introduced by the positional-wise sMLP operation and $\odot$ represents the element-wise multiplication.

The increase of parameters for the sMLP is $d (d + 1)$, where $d$ is the embedding dimension (e.g., 256). 
The ReLU activation is necessary to change the feature distributions of the foreground tokens so that the foreground tokens always have non-negative values to encode the foreground/background information.

\noindent\textbf{Outputs of triple-attention.} 
With the three groups of $(Q, K, V)$, we can generate the outputs $Z^{Attn_1}$, $Z^{Attn_2}$ and $Z^{Attn_3}$ for $Attn1$, $Attn2$ and $Attn3$, respectively, by sharing the subsequent modules (e.g., $\mha$ and $\ffn$) in the same encoder layer with Eq. \ref{eq:encoder-forward}.
The workflow of how to obtain $Z^{Attn_1}$, $Z^{Attn_2}$ and $Z^{Attn_3}$ is shown in Fig. \ref{fig:triple-attention-module}.  
 These outputs are further processed separately by the shared subsequent decoder layers to keep the training of each attention independent. The second and third attentions as well as the sMLP are removed during inference so no additional parameters and computation overhead are introduced after training.



\noindent\textbf{Loss function.} Our loss function is a combination of the default loss of the original DETR applied to the predictions using the outputs $Z^{Attn_1}$, $Z^{Attn_2}$ and $Z^{Attn_3}$ of the triple-attentions: 

\begin{equation}\label{eq:loss} 
 L = L_{det}^{Attn_1} + L_{det}^{Attn_2} + L_{det}^{Attn_3}.
\end{equation}


\begin{table*}[t!]
 \centering
\caption{Results of DETR baseline methods and our KS-DETR. } 
\label{tab:baseline}
\begin{tabular}{llllllll}  
\hline
Model          & \#Epochs & $\AP$         & $\AP_{50}$  & $\AP_{75}$   & $\AP_{S}$   & $\AP_{M}$      & $\AP_{L}$  \\
\hline
Conditional-DETR-R50   &   50       & 41.3        & 62.5 & 43.6 & 21.0 & 44.6 & 59.6 \\
KS-Conditional-DETR-R50&   50       & 42.1 (+0.8) & 63.4 & 44.8 & 21.4 & 45.9 & 60.6 \\
Conditional-DETR-R101 \cite{meng2021conditional}  & 50       & 42.8        & 63.7 & 46.0 & 21.7 & 46.6 & 60.9 \\
KS-Conditional-DETR-R101     & 50       & 43.4 (+0.6) & 64.8 & 46.7 & 23.5 & 47.2 & 62.3 \\
\hline
DAB-DETR-R50           &   50       & 43.0        & 63.5 & 45.8 & 22.4 & 46.6 & 61.3 \\
KS-DAB-DETR-R50        &   50       & 43.9 (+0.9) & 64.2 & 46.8 & 23.9 & 48.0 & 62.2 \\
DAB-DETR-R101  \cite{ideacvr2022detrex}      &   50       & 44.0       & 62.9 & 47.6 & 23.8 & 48.4 & 61.8 \\
KS-DAB-DETR-R101       &   50    & 45.3 (+1.3) & 65.4 & 48.8 & 24.3 & 49.6 & 63.6 \\
DAB-DETR-Swin-T \cite{ideacvr2022detrex}  & 50  & 45.2  & 66.8 & 47.8 & 24.2 & 49.0 & 64.8 \\
KS-DAB-DETR-Swin-T     &    50 & 47.1 (+1.9)  & 68.3    & 50.2  & 27.1 & 51.3 & 66.5  \\
\hline
DN-DETR-R50            &   50     & 44.7        & 64.8 & 47.5 & 23.4 & 48.9 & 63.7 \\
KS-DN-DETR-R50         &  50       & 45.2 (+0.5) & 	64.9 & 48.2 & 24.4 & 49.3 & 63.0 \\
DN-DETR-R101           &   50       & 45.6        & 65.9 & 49.0 & 24.4 & 49.9 & 64.0 \\
KS-DN-DETR-R101        &   50       & 46.5 (+0.9) & 66.2 & 49.8 & 25.7 & 50.8 & 65.2 \\
\hline
Other Multi-scale DETR variants &      &             &      &      &      &      &      \\
\hline
Deformable-DETR-R50     & 12       & 35.3        & 51.8 & 38.2 & 19.2 & 39.1 & 47.2 \\
KS-Deformable-DETR-R50 & 12       & 36.4 (+1.1) & 53.5 & 39.5 & 20.1 & 39.5 & 48.2 \\
Deformable-DETR-R101    & 12       & 36.8        & 54.2 & 40.0 & 21.1 & 40.3 & 49.2 \\
KS-Deformable-DETR-R101 & 12       & 38.4 (+1.6) & 55.9 & 41.8 & 21.5 & 42.3 & 51.6 \\
DN-Deformable-DETR-R50 \cite{li2022dn}  & 12       & 43.4        & 61.9 & 47.2 & 24.8 & 46.8 & 59.4 \\
KS-DN-Deformable-DETR-R50   & 12       & 46.5 (+2.1) & 63.9 & 50.4 & 28.8 & 49.5 & 61.5 \\
\hline
Deformable-DETR-R101    & 24       & 41.6        & 59.6 & 45.3 & 24.3 & 45.2 & 55.6 \\
KS-Deformable-DETR-R101 & 24       & 43.0 (+1.4) & 61.1 & 47.1 & 24.9 & 46.6 & 57.0 \\
\hline
Deformable-DETR-R50 \cite{zhu2020deformable}     & 50       & 44.1        & 62.6 & 47.7 & 26.4 & 47.1 & 58.0 \\
KS-Deformable-DETR-R50  & 50       & 44.8 (+0.7) & 62.9 & 48.7 & 26.9 & 48.4 & 58.9 \\
Deformable-DETR-R101    & 50       & 45.1        & 63.5 & 49.1 & 27.4 & 48.8 & 59.9 \\
KS-Deformable-DETR-R101 & 50       & 46.0 (+0.9) & 64.3 & 50.1 & 28.9 & 49.7 & 60.3 \\
\hline
\end{tabular}
\end{table*}

\section{Experiments}
\label{sec:experiment}

\begin{table*}[t!]
 \centering
\caption{Ablation results of our KS-DETR built upon DAB-DETR-R50 and DAB-DETR-R101. All the models are trained with 50 epochs.} 
\label{tab:effect-triple-attention}

\begin{tabular}{llllllllll}
\hline
Model  & Exp.  & Sharing $V$ & Sharing $A$ &  $\AP$  & $\AP_{50}$  & $\AP_{75}$   & $\AP_{S}$   & $\AP_{M}$  & $\AP_{L}$    \\
\hline

\multirow{4}{*}{DAB-DETR-R50}  & Baseline         &     &     & 43.0 & 63.5 & 45.8 & 22.4 & 46.6 & 61.3 \\
                               & Dual-attention   & \checkmark &     & 43.4 & 63.5 & 46.0 & 23.4 & 47.2 & 61.3 \\
                               & Dual-attention   &     & \checkmark & 43.6 & 63.8 & 46.4 & 24.6 & 47.6 & 61.8 \\
                               & Triple-attention & \checkmark & \checkmark & 43.9 & 64.2 & 46.8 & 23.9 & 48.0 & 62.2 \\
\hline
\multirow{4}{*}{DAB-DETR-R101} & Baseline         &     &     & 44.0 & 62.9 & 47.6 & 23.8 & 48.4 & 61.8 \\
                               & Dual-attention   & \checkmark &     & 45.0 & 65.3 & 47.8 & 25.6 & 48.9 & 63.3 \\
                               & Dual-attention   &     & \checkmark & 45.2 & 65.4 & 48.4 & 25.9 & 49.4 & 62.7 \\
                               & Triple-attention & \checkmark & \checkmark & 45.3 & 65.4 & 48.8 & 24.3 & 49.6 & 63.6 \\
\hline
\end{tabular}
\end{table*}

 \subsection{Main results}
\label{sec:exp-baselines}

We use the COCO dataset \cite{lin2014microsoft} to evaluate our method and report the AP on the COCO 2017 validation set. 
We compare our method with the following baseline methods: Conditional-DETR \cite{meng2021conditional}, DAB-DETR \cite{liu2022dab}, DN-DETR \cite{li2022dn}, Deformable-DETR \cite{zhu2020deformable} and DN-Deformable-DETR \cite{li2022dn}.
Among them, Deformable-DETR \cite{zhu2020deformable} and DN-Deformable-DETR \cite{li2022dn} use multi-scale features, and the rest use single-scale features. 
We use ResNet50 (R50 \cite{he2016deep}) and ResNet101 (R101 \cite{he2016deep}) as the backbones. For DAB-DETR \cite{liu2022dab}, we further test the transformer backbone Swin-T \cite{liu2021swin}. 
We follow the standard training procedures of DETR for the 12-epoch and 50-epoch training with batch size 16. 

The results are shown in Table \ref{tab:baseline}. We see that our method consistently improves all baseline methods for all tested backbones. There are two interesting patterns. 
First, the improvement for a powerful backbone tends to be relatively large than a weak backbone. Take DAB-DETR \cite{liu2022dab} as an example, the improvements over the baseline are 1.9, 1.3 and 0.6 AP for Swin-T, R101 and R50, respectively. 
Second, the training with short training schedules (e.g., 12 epochs) generally exhibits larger improvements than that with long training schedules (e.g., 50 epochs), suggesting that our method speeds up the training. For instance, the improvements for Deformable-DETR-R101 \cite{zhu2020deformable} at 12, 24 and 50 epochs are 1.6, 1.4 and 0.9 AP, respectively.
 The fast convergence of our method can also be seen in Fig. \ref{fig:teacher-attn-accu} by comparing the baseline with $Attn1$ (the plain attention in our triple-attention module).



\subsection{Effects of GT Fg-Bg Mask on learning of teacher attention and values}

\begin{figure}[t!]
\centering
\subfigure{\includegraphics[width=0.45\textwidth]{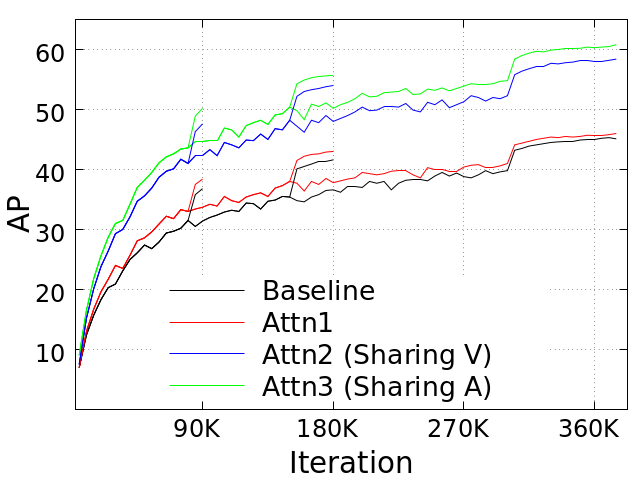}}
\caption{Detection results with the outputs of each attention in our triple-attention module for KS-Deformable-DETR-R101. Both the baseline and our method are trained for 50 epochs (375K iterations). Note that $Attn2$ and $Attn3$ use the GT Fg-Bg Mask during both training and inference, and $Attn1$ and the baseline never access the GT Fg-Bg Mask.}
\label{fig:teacher-attn-accu}
\end{figure}

The design of our method is to first learn high-quality $A$ and $V$ and then drive the shared $V$ and $A$ to a higher level of quality. Here we verify if we have learned high-quality $A$ and $V$ for the two teacher attentions. 
We plot the accuracies of the predictions for each attention ($Attn1$, $Attn2$ and $Attn3$) of our triple-attention in Fig. \ref{fig:teacher-attn-accu}. We see that the two teacher attentions ($Attn2$ and $Attn3$) outperform the student attention $Attn1$ with a large margin (10+ AP) with the assistance of GT Fg-Bg Mask. 

All the parameters of the model used for making predictions for $Attn1$ and $Attn2$ are shared, except the ones used for estimating their attention maps. Thus it is clear that $Attn2$ has learned the much higher quality of attention maps compared with $Attn1$. If we replace the attention map of $Attn1$ with that of $Attn2$, we can immediately obtain the same high accuracy of $Attn2$. 
Similarly, we confirm that high-quality values are learned for the third attention by comparing the accuracies of $Attn1$ and $Attn3$.

\subsection{Effects of knowledge-sharing in triple-attention}
\label{sec:ablation-effect-of-sharing}

 We verify the effectiveness of the knowledge-sharing strategy in our second and third attentions. 
We compare our triple-attention with two dual-attention modules: 1) a plain attention and a second attention with high-quality values (with the assistance of GT Fg-Bg Mask) but sharing $A$ with the plain attention (Fig. \ref{fig:knowledge-sharing-framework}b), and 2) a plain attention and a second attention with high-quality attention maps but sharing $V$ with the plain attention (Fig. \ref{fig:knowledge-sharing-framework}c). 

We use DAB-DETR-R50 and DAB-DETR-R101 as the baseline methods. The results are shown in Table. \ref{tab:effect-triple-attention}. We see that dual-attentions (sharing $V$ or sharing $A$) outperform the baseline (single attention) and triple-attention exhibits the biggest improvements. 
The experiments demonstrate the effectiveness of the strategy of sharing $V$ and $A$ in attention learning and justify our design of triple-attention. However, the improvements for triple-attention over dual-attentions are tiny, this is an issue we will address in our future work.

\section{Conclusions}
\label{sec:conclusion}

In this paper, we propose a triple-attention module to improve the learning of scaled dot-product attention in the detection transformer. 
We use GT Fg-Bg Mask as additional cues to learn good teacher attention maps and values to eliminate the need of training a large teacher model.
 We design two teacher attentions to improve the learning of the attention maps and values of the plain student attention by sharing attention maps and values.  
 Our method exhibits consistent improvements over various DETR-like baseline methods.

{\small
\bibliographystyle{ieee_fullname}
\bibliography{egbib}
}

\end{document}